\documentclass{article}
\usepackage{ijcai24}

\usepackage{amsmath}
\usepackage{mathtools}

\usepackage{times}
\usepackage{soul}
\usepackage{url}
\usepackage[hidelinks]{hyperref}
\usepackage[utf8]{inputenc}
\usepackage[small]{caption}
\usepackage{graphicx}
\usepackage{amsmath}
\usepackage{amsthm}
\usepackage{booktabs}
\usepackage{algorithm}
\usepackage{algorithmic}
\usepackage[switch]{lineno}

\usepackage{booktabs} 
\usepackage{array}
\usepackage{multirow}

\usepackage{subcaption} 
\usepackage{makecell}
\usepackage{threeparttable}

\title{EDGE: Efficient Data Selection for LLM Agents via Guideline Effectiveness}

\begin{document}


\author{
Yunxiao Zhang$^1$
\and
Guanming Xiong$^1$\and
Haochen Li$^{2}$\And
Wen Zhao$^1$\thanks{Corresponding author.}    \\
\affiliations
$^1$Peking University\\
$^2$01.AI\\
\emails
yunxiao.zhang@stu.pku.edu.cn,
zhaowen@pku.edu.cn
}

\maketitle

\begin{abstract}
    Large Language Models (LLMs) have shown remarkable capabilities as AI agents. However, existing methods for enhancing LLM-agent abilities often lack a focus on data quality, leading to inefficiencies and suboptimal results in both fine-tuning and prompt engineering. To address this issue, we introduce EDGE, a novel approach for identifying informative samples without needing golden answers. We propose the Guideline Effectiveness (GE) metric, which selects challenging samples by measuring the impact of human-provided guidelines in multi-turn interaction tasks. A low GE score indicates that the human expertise required for a sample is missing from the guideline, making the sample more informative. By selecting samples with low GE scores, we can improve the efficiency and outcomes of both prompt engineering and fine-tuning processes for LLMs. Extensive experiments validate the performance of our method. Our method achieves competitive results on the HotpotQA and WebShop and datasets, requiring 75\% and 50\% less data, respectively, while outperforming existing methods. We also provide a fresh perspective on the data quality of LLM-agent fine-tuning.
\end{abstract}


\section{Introduction}

In recent years, Large Language Models (LLMs) \cite{Ouyang-Long-NeurIPS-2022-InstructGPT,OpenAI-2023-GPT4} have demonstrated remarkable few-shot learning and reasoning capabilities. An increasing number of studies have begun exploring how to leverage LLMs as agents that can accomplish various tasks through multiple interactions with the environment \cite{deng-2023-mindweb,liu-2024-ICLR-agentbench,Wang-2024-agentsurvey}. For example, WebShop \cite{Yao-Shunyu-NeurIPS-2022-WebShop} provides a simulated shopping environment where agents must select products that best match user requirements.

During interactions, LLMs frequently encounter complex or previously unseen scenarios, which places substantial demands on their generalization capabilities. Numerous studies have been dedicated to mitigating this challenge.

%

Prior work has demonstrated the importance of guidelines (or insights) in prompt-based multi-turn interaction methods. 
Guidelines are natural language prompts summarized from data that contain more information and cover more scenarios than exemplars, while typically consuming less context space. 
Existing approaches autonomously gather experiences from training tasks through trial and error to generate these guidelines \cite{Zhao-Andrew-AAAI-2024-ExpeL}.

Another line of research focuses on Supervised Fine-Tuning (SFT) of open-source LLMs to enhance their instruction-following capabilities. Prior work has shown that the effectiveness of SFT depends more on dataset quality than quantity \cite{wang-etal-2023-self-instruct,Zhou-Chunting-NeurIPS-2023-LIMA}. Current data filtering approaches, including GPT-4-based scoring \cite{Chen-Lichang-ICLR-2024-AlpaGasus}, instruction difficulty assessment \cite{Li-Ming-NAACL-2024-IFD}, and semantic diversity metrics \cite{lu-keming-ICLR-2024instag}, have demonstrated varying degrees of success.

Despite these advancements, current LLM-agent approaches still face several pressing challenges. In prompt-based methods, existing approaches for obtaining guidelines \textbf{do not consider data quality control}, instead randomly selecting samples from annotated data, which not only requires substantial and costly annotation efforts but also suffers from noisy data problems. Meanwhile, in SFT-based methods, current approaches heavily \textbf{rely on golden answer feedback} and primarily focus on single-turn instruction tuning, lacking necessary exploration of more complex multi-turn interaction scenarios that are essential for real-world applications.

To address these challenges, we propose \underline{E}fficient \underline{D}ata selection for LLM agents via \underline{G}uideline \underline{E}ffectiveness, a novel framework centered around a new metric called Guideline Effectiveness (GE) to select the most informative subset of samples from a vast unlabeled data (query) pool. These selected samples can be utilized for both prompt engineering and SFT. 

Guidelines represent human understanding of tasks and serve as prior knowledge for agents, encompassing tool usage patterns and comprehension of complex scenarios \cite{Zhao-Andrew-AAAI-2024-ExpeL,Fu-Yao-NeurIPS-2024-AutoGuide}. The GE score essentially quantifies the impact of guidelines on each data sample, enabling us to identify which samples are most challenging for the model and thus select more informative ones.
Beginning with an initial guideline, we employ an active learning approach to select a small number of samples with the lowest GE scores. These samples are then analyzed to summarize error causes and update the guideline. Next, we use the updated guideline and advanced API-based LLM to annotate more low-GE-score samples instead of relying on human annotators. Notably, the updated guideline incorporates solutions for challenging samples and deeper insights into the task and tools, ensuring that the annotated data is of high quality. Finally, we can use these informative and high-quality annotated samples to fine-tune open-source LLMs.

The main contributions of this work are summarized as follows:
\begin{itemize}
    \item Propose a novel Guideline Effectiveness metric to identify informative samples using guidelines without requiring golden answers. This metric enables efficient sample selection for both prompt engineering and model fine-tuning.

    
    \item Derive effective guidelines and obtain high-quality data for challenging multi-turn interaction tasks without the need for manual annotation, by leveraging the GE score.
    
    \item Demonstrate the effectiveness of our approach through extensive experiments on HotpotQA and WebShop benchmarks, achieving state-of-the-art performance with only 75\% and 50\% data requirements compared to existing methods.
\end{itemize}

\section{Related Work}

This study investigates how to effectively utilize guidelines in the context of data selection for supervised fine-tuning (SFT).

\textbf{Data Selection for SFT} aims to select a high-quality subset of data. 
\cite{Zhou-Chunting-NeurIPS-2023-LIMA} demonstrates that only 1,000 carefully curated prompts and responses can achieve remarkably strong performance.
\cite{Chen-Lichang-ICLR-2024-AlpaGasus} proposes using GPT-4 for direct quality scoring, successfully identifying 9k high-quality samples from a dataset of 52k instances.
\cite{Li-Ming-NAACL-2024-IFD} introduces the Instruction-Following Difficulty (IFD) metric to identify discrepancies between a model's expected responses and its intrinsic generation capabilities.
\cite{Liu-Wei-ICLR-2024-DEITA} curates 6K training samples by evaluating them along three dimensions: complexity, quality, and diversity.
\cite{Bhatt-Gantavya-ACL-2024-ExperimentalDesign} conducts a comprehensive evaluation of existing data selection methods that aim to maximize uncertainty and/or diversity measures.
However, these evaluation metrics inherently depend on golden answers as feedback. Furthermore, they primarily focus on single-turn interactions, neglecting the complexities of multi-turn interaction scenarios.
AgentTuning \cite{zeng-etal-2024-agenttuning} and FiReAct \cite{Chen-Baian-2023-FireAct} investigate fine-tuning LLMs with multi-turn interaction trajectories generated by GPT-4, further examining the effects of multi-task learning and prompt design methods, respectively.
However, both methods randomly select samples for annotation, and assume that perfectly correct trajectories (reward = 1) represent high quality. This approach may result in the inclusion of simpler problems in fine-tuning datasets, leading to low quality of fine-tuning data.

\textbf{Deep Active Learning} aims to identify the most informative samples for annotation, thereby reducing labeling costs. The methods are typically categorized into uncertainty-based \cite{Settles-Burr-JMLR-2011-ActiveLearning,Kremer-Jan-WIREs-DMKD-2014-ActiveSVM}, diversity-based \cite{Sener-Ozan-ICLR-2018-CoreSet,bukharin-etal-2024-data-Diversity,Bhatt-Gantavya-ACL-2024-ExperimentalDesign}, or hybrid approaches \cite{azeemi-etal-2025-label}. In the era of large language models (LLMs), some studies have attempted to integrate active learning with LLMs to achieve efficient SFT. \cite{azeemi-etal-2025-label} investigates active learning for improving label efficiency in natural language generation but reports inconsistent results. \cite{Kung-Po-Nien-EMNLP-2024-ActiveInstruction} proposed a task-level active learning framework to explore the most effective SFT tasks. However, it makes the simplifying assumption that all instances are of equal value within a task.\cite{Bhatt-Gantavya-ACL-2024-ExperimentalDesign}is most similar to ours. It is the first to utilize experimental design for SFT and formulates active learning as a facility location problem. This method focuses on selecting semantically diverse and representative samples, effectively improving the generative capabilities of LLMs. However, it does not focus on addressing agent tasks that require more reasoning and decision-making capabilities.

\textbf{Guideline-based Prompting} aims to summarize historical interaction experiences from datasets into natural language prompts that can guide future interactions. 
\cite{Zhao-Andrew-AAAI-2024-ExpeL} introduces Experiential Learning, which autonomously gathers experiences from training tasks through trial and error to generate instructive guidelines. 
\cite{Fu-Yao-NeurIPS-2024-AutoGuide} advances this approach by automatically generating context-aware guidelines and implementing a retrieval system that selects guidelines relevant to the agent's current state.
However, these approaches rely on random sampling without quality consideration and their automated summarization lacks the depth and nuance of expert knowledge.

\section{Methodology}

\begin{figure*}[ht]
    \centering
    \includegraphics[page=10, width=1\textwidth]{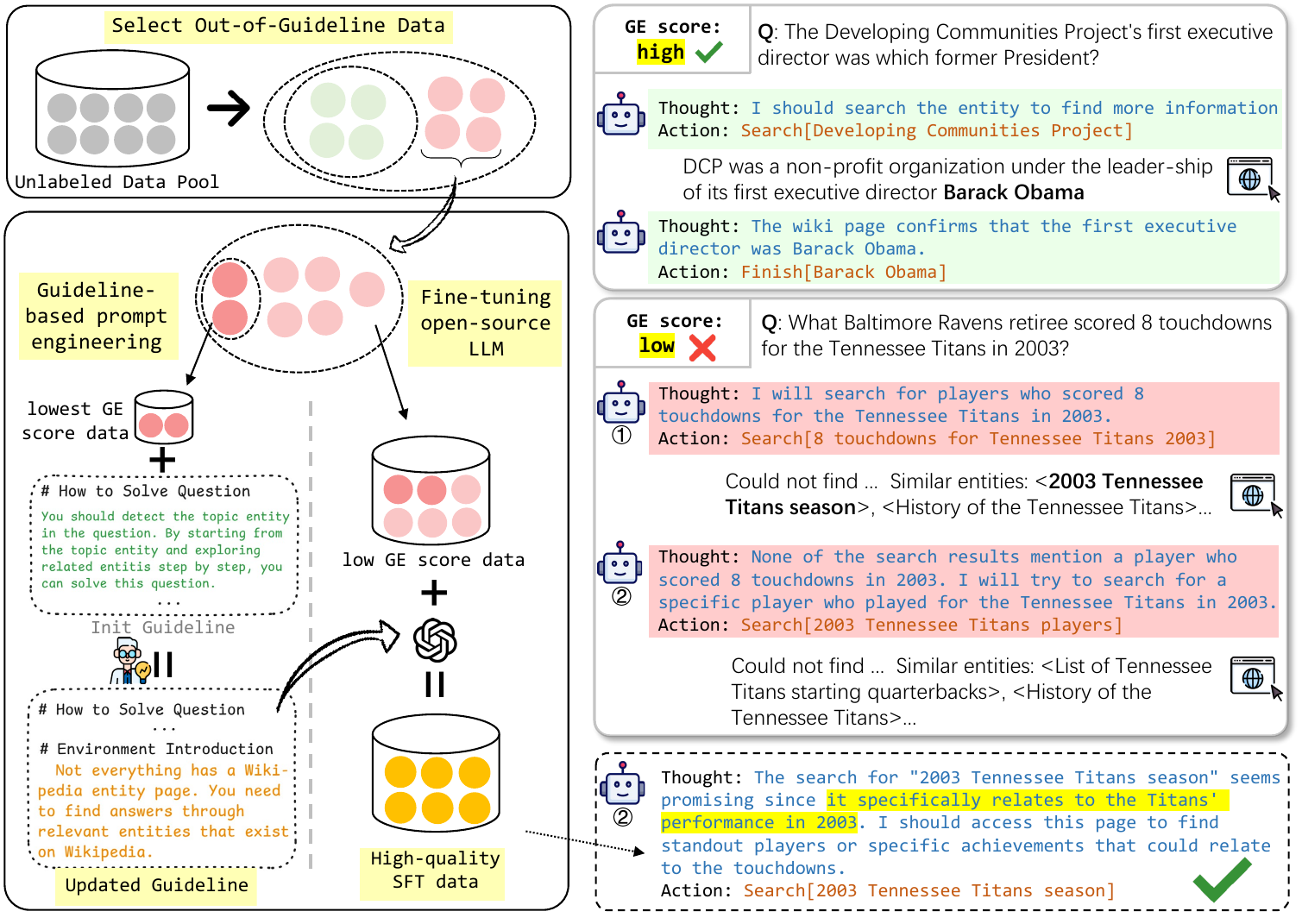}
    \caption{The overall process of our method and an example of how guidelines correct LLM's behavior}
    \label{fig:main}
\end{figure*}

\subsection{Overview}

Our core insight is to identify informative samples from an unlabeled data pool by leveraging guidelines, as illustrated in Figure \ref{fig:main}.
Given an unlabeled data pool and initial guidelines, we first compute the Guideline Effectiveness (GE) score for each sample using the initial guidelines. Samples with lower scores are selected for manual annotation, resulting in updated guidelines that can be directly applied to prompt-based methods.
To fully utilize the unlabeled data, we incorporate the new guidelines into the prompt text and employ GPT-4 for annotation, generating question-interaction trajectory pairs as high-quality SFT data. Notably, this entire process does not require golden answers.

\subsection{Preliminary}

Given a set of questions $\mathcal{Q}=\{q_1,\dots,q_n\}$, a language model LLM, and an initial guideline $\mathcal{G}^{init}$, we generate interaction trajectories $\mathcal{T}=\text{LLM}(\mathcal{Q}, \mathcal{G}^{init})$, where each trajectory $\mathcal{T}=(q, a_1, o_1, \ldots, a_T,o_T)$ consists of question-action-observation sequences with length $T$. Here $a_i$ represents the action taken by the LLM at step $i$, and $o_i$ denotes the observation or feedback received from the environment after taking action $a_i$.

Our objective is to select an informative subset $\mathcal{Q}'$ such that:

\begin{equation}
    \mathcal{Q}' = \mathop{\arg\max}_{\mathcal{Q}' \subset \mathcal{Q}} (\text{Reward}(\text{LLM'}))
\end{equation}
where Reward is the metrics of the task and LLM' is fine-tuned by labeled $\mathcal{Q}'$.


\subsection{Guideline Effectiveness}

Guidelines are natural language prompts enriched with expert knowledge that can cover more scenarios while consuming less context space compared to detailed exemplars. 
However, since humans may not initially recognize all potential challenging samples, the initial guidelines may be inadequate.
Therefore, we propose a metric called Guideline Effectiveness to quantify the contribution of guidelines in solving a given question in order to identify questions that are challenging for the initial guidelines.


Given an question $q$, the prompt text is constructed as:
\begin{equation}
    \text{Prompt}=\text{CONCAT}(\text{I}, \text{G}, \text{E}) 
\end{equation}

where $\text{I}$ represents the instruction text, and $\text{E}$ denotes a set of interaction examplars. 
We measure the uncertainty of LLM's output action $a_t$ at step $t$ using the average cross-entropy loss of each token.

\begin{equation}
d_\theta^G(a_t|\text{Prompt}) = -\frac{1}{N}\sum_{i=1}^{N} \log P(w_i | \text{I},\text{G}, \text{E}, \mathcal{T}, w_{<i}:\theta)
\end{equation}

where $N$ is the number of tokens in $a_t$, $w_i$ is the $i$-th token in $a_t$. A lower $d_\theta^G$ indicates easier action generation.

To evaluate guideline effectiveness, we construct $\text{Prompt}^{-G} = \text{CONCAT}(\text{I}, \text{E})$ by excluding $\text{G}$ from the context. The difficulty of generating action without guidelines is:

\begin{equation}
d_\theta^I(a_t|\text{Prompt}^{-G}) = -\frac{1}{N}\sum_{i=1}^{N} \log P(w_i | \text{I},\text{E}, \mathcal{T}, w_{<i}:\theta)
\end{equation}

The score $d_\theta^I$ measures how hard it is to generate $a$ using only the LLM's intrinsic knowledge, without guidelines.

Based on $d_\theta^G$ and $d_\theta^I$, the GE score is defined as:   

\begin{equation}
GE(q) = -\frac{1}{T}\sum_{t=1}^T \log\frac{d_\theta^I(a_t|\text{Prompt}^{-G})}{d_\theta^G(a_t|\text{Prompt})}
\end{equation}

GE quantifies the influence of guideline on generating each $a_t$ by computing the ratio between the difficulty scores with and without guidelines. 
The intuitive interpretation of GE values is as follows:
A lower value of $d$ indicates the generation process is easier. The difficulty scores $d_\theta^{I}$ and $d_\theta^{G}$ represent generation difficulty without and with guidelines respectively.
When \textbf{$GE > 0$}, we have $d_\theta^{I} > d_\theta^{G}$, indicating that guidelines facilitate action generation. A larger positive GE score suggests guidelines have a stronger positive impact on generation. As the score approaches zero, the similar magnitudes of $d_\theta^{I}$ and $d_\theta^{G}$ indicate guidelines provide limited benefit - these are samples of particular interest.
When \textbf{$GE < 0$}, meaning $d_\theta^{I} < d_\theta^{G}$, guidelines appear to impede generation. This reveals cases where the LLM's inherent knowledge leads it to generate actions that conflict with the guidelines, another important category of samples to identify.


\subsection{Efficiently Incorporating Human Expertise}

This section describes how we utilize the GE score to incorporate human expertise into LLMs efficiently, as shown in Figure \ref{fig:main}. Given the data pool $\mathcal{Q}$ and historical interaction trajectory $\mathcal{T}$, and initial guideline $\mathcal{G}^{init}$, we can calculate the GE score for each $q_i$. Lower GE scores indicate challenging questions which require additional learning by the LLM. We accomplish this in two stages: Guideline Update and High-Quality Data Generation.

\textbf{Update the Guideline.} We select m questions with the lowest GE scores. By observing these questions' interaction trajectories, we can summarize their issues and update $\mathcal{G}^{init}$ to $\mathcal{G}^{new}$. Analyzing samples with the lowest GE scores allows $\mathcal{G}^{new}$ to integrate further human expertise necessary for addressing challenging samples, such as deeper insights into the task and tools. The value of m can be small (e.g., 30), which is manageable by humans within a reasonably short time. We denote the ReAct framework augmented with the updated guideline as $\text{EDGE}_\text{UG}$.

\textbf{High-Quality Data Generation.} Similarly, we select k questions with the lowest GE scores and employ GPT-4 to generate interaction trajectories guided by $\mathcal{G}^{new}$. The incorporation of human expertise within $\mathcal{G}^{new}$ ensures that the trajectories  maintain a high standard of quality. Utilizing these high-quality data for fine-tuning, open-sourced LLMs will implicitly learn human expertise from the annotated data. Fine-tuning the open-sourced LLM is often crucial because: 1) As guidelines become more complex, it gets harder for open-sourced LLM to follow; 2) Some tasks may be too intricate to distill into guidelines, making annotating data a simpler option.

\section{Experiments}

\subsection{Baselines}
To evaluate our approach, we have selected a range of state-of-the-art (SOTA) methods as baselines and conducted model comparisons along two dimensions: 

\textbf{$\text{EDGE}$ vs. Other Agent Methods}. We compare our method against several state-of-the-art agent methods: \textbf{ReAct} \cite{Yao-Shunyu-ICLR-2023-ReAct} integrates reasoning and acting capabilities for sequential decision-making tasks; \textbf{Reflexion} \cite{Shinn-Noah-NeurIPS-2023-Reflexion} reinforces language agents through linguistic feedback; \textbf{AMOR} \cite{Guan-Jian-NeurIPS-2024-AMOR} constructs reasoning logic over finite state machines for automated problem-solving across modules; \textbf{ExpeL} \cite{Zhao-Andrew-AAAI-2024-ExpeL} leverages GPT-4 to extract guidelines from failed trajectories.

\textbf{$\text{GE}$ vs. Other Data Selection Strategies.} For comparison with other label-efficient data selection strategies, we evaluate GE against several baseline approaches: \textbf{Random} selects data randomly for annotation; \textbf{Mean Entropy} \cite{Settles-Burr-JMLR-2011-ActiveLearning,Kremer-Jan-WIREs-DMKD-2014-ActiveSVM} measures uncertainty through token-wise negative entropy of softmax probabilities; \textbf{FL} \cite{Bhatt-Gantavya-ACL-2024-ExperimentalDesign} selects semantically representative samples based on diversity; and \textbf{High Score} \cite{Chen-Baian-2023-FireAct,zeng-etal-2024-agenttuning} retains only fully correct interaction trajectories from annotated data.

\subsection{Experiment Setup}

\textbf{Datasets.} \textbf{WebShop} \cite{Yao-Shunyu-NeurIPS-2022-WebShop} is a simulated online shopping environment composed of a website with 1.18M real-world products. The agent's goal is to purchase a product that meets specific requirements based on a text instruction. This task requires the agent to query the website’s search engine, select products with required features, and click the necessary options. following \cite{liu-2024-ICLR-agentbench}, the system implements two valid actions: search[query] and click[button]. \textbf{HotpotQA} \cite{yang-etal-2018-hotpotqa} is a multi-hop question-answering benchmark that challenges an agent to retrieve Wikipedia passages to perform reasoning and question-answering. This involves utilizing API calls and LLM's knowledge to search for and retrieve information in order to find answers. Following \cite{Yao-Shunyu-ICLR-2023-ReAct}, we use three types of actions to support interactive information retrieval in HotpotQA: search[entity], lookup[query] and finish[answer].

For Webshop, we use 8,500 instructions as the data pool and another 500 instructions for evaluation. For HotpotQA, we use the first 10,000 training questions as the data pool and randomly select 500 dev questions. For each dataset, we selected 30 samples with the lowest GE score for guideline updating, and then annotated 800 samples for fine-tuning. The statistical details of the test datasets are presented in Table \ref{tab:dataset-Statistics}.

\begin{table}[h]
  \centering
  \begin{tabular}{lccc}
    \toprule
    \textbf{Dataset} & \textbf{\#Data Pool} & \textbf{\makecell{\#EDGE Used \\  (m / k)}} & \textbf{\makecell{\#Raw \\ (Train / Dev)}}  \\
    \midrule
    WebShop & 8,500 & 30 / 800 & 12,087 / - \\
    HotpotQA & 10,000 & 30 / 800 & 90,564 / 7405 \\
    \bottomrule
  \end{tabular}
  \caption{Statistics of the datasets.}
  \label{tab:dataset-Statistics}
\end{table}

\begin{figure*}[ht]
    \centering
    \begin{minipage}{0.56\textwidth}
    \captionsetup{type=table} 
        \centering
        \begin{tabular}{@{}lccccc@{}} 
            \toprule
            \multirow{2}{*}{\textbf{Method}} & \multirow{2}{*}{\textbf{Base Model}} & \multicolumn{2}{c}{\textbf{HotpotQA}} & \multicolumn{2}{c}{\textbf{WebShop}} \\
            \cmidrule(r){3-4} \cmidrule(l){5-6}
             &    & \textbf{EM} & \textbf{F1} & \textbf{Reward}  & \textbf{SR}\\ 
            \midrule
            \multicolumn{6}{c}{\textbf{API-based LLM}} \\ 
            \midrule
            ReAct  & GPT-4o     & 42.0& 55.17     & 58.63     & 33.2     \\
            ExpeL  & GPT-4o     & 47.8& 60.92     & \underline{64.16}     & \underline{42.0}     \\
            Reflexion& GPT-4o     & 49.2& 61.30     & 63.28     & 39.8     \\
            AMOR\textsuperscript{\dag}   & GPT-4-turbo& \underline{55.2}& \underline{65.20}     & -& -\\
            \midrule
            $\text{EDGE}_\text{UG}$ (Ours) & GPT-4o     & \textbf{63.7}     & \textbf{72.88}    & \textbf{73.11}   & \textbf{47.8}    \\
            \midrule
            \multicolumn{6}{c}{\textbf{Fine-tuned Open-source LLM}}\\ 
            
            \midrule
            
            
            ReAct       & M-7B & 22.6     & 38.31     & 30.77    & 14.2     \\
            \quad w/ Random   & M-7B & 34.4& 46.11     & 59.05    & 39.0     \\
            \quad w/ Mean Entropy& M-7B & 35.8& 47.00     & 58.79     & 38.8     \\
            \quad w/ High Score  & M-7B & \underline{37.2}& \underline{49.19}     & \underline{59.32}    & \underline{39.2}     \\
            \quad w/ FL  & M-7B & 32.8& 46.06     & 59.00    & 39.0     \\
            \midrule
            \quad w/ GE (Ours)   & M-7B & \textbf{41.8}     & \textbf{55.47}    & \textbf{62.07}   & \textbf{41.2}    \\
            \midrule
            
            ReAct       & L-8B & 35.4     & 45.96     & 37.42    & 18.0     \\
            \quad w/ Random   & L-8B & 44.2& 56.02     & \underline{66.73}    & 42.8     \\
            \quad w/ Mean Entropy& L-8B & 46.0& 56.13     & 64.3     & 42.4     \\
            \quad w/ High Score  & L-8B & \underline{46.6}& \underline{57.66}     & 66.21    & \underline{43.6}     \\
            \quad w/ FL  & L-8B & 40.6& 52.38     & 66.09    & 43.2     \\
            \midrule
            \quad w/ GE (Ours)   & L-8B & \textbf{52.4}     & \textbf{66.15}    & \textbf{69.14}   & \textbf{46.0}    \\
            
            \bottomrule
        \end{tabular}
        \caption{Main results. The best results are marked in \textbf{bold} and the second-best results are marked with \underline{underline}. Results marked with \textsuperscript{\dag} are reported in the original paper.}
        \label{tab:maintab}

    \end{minipage}%
    \hspace{0.5cm} 
    \begin{minipage}{0.38\textwidth}
        \centering
        \includegraphics[page=1, width=0.9\textwidth]{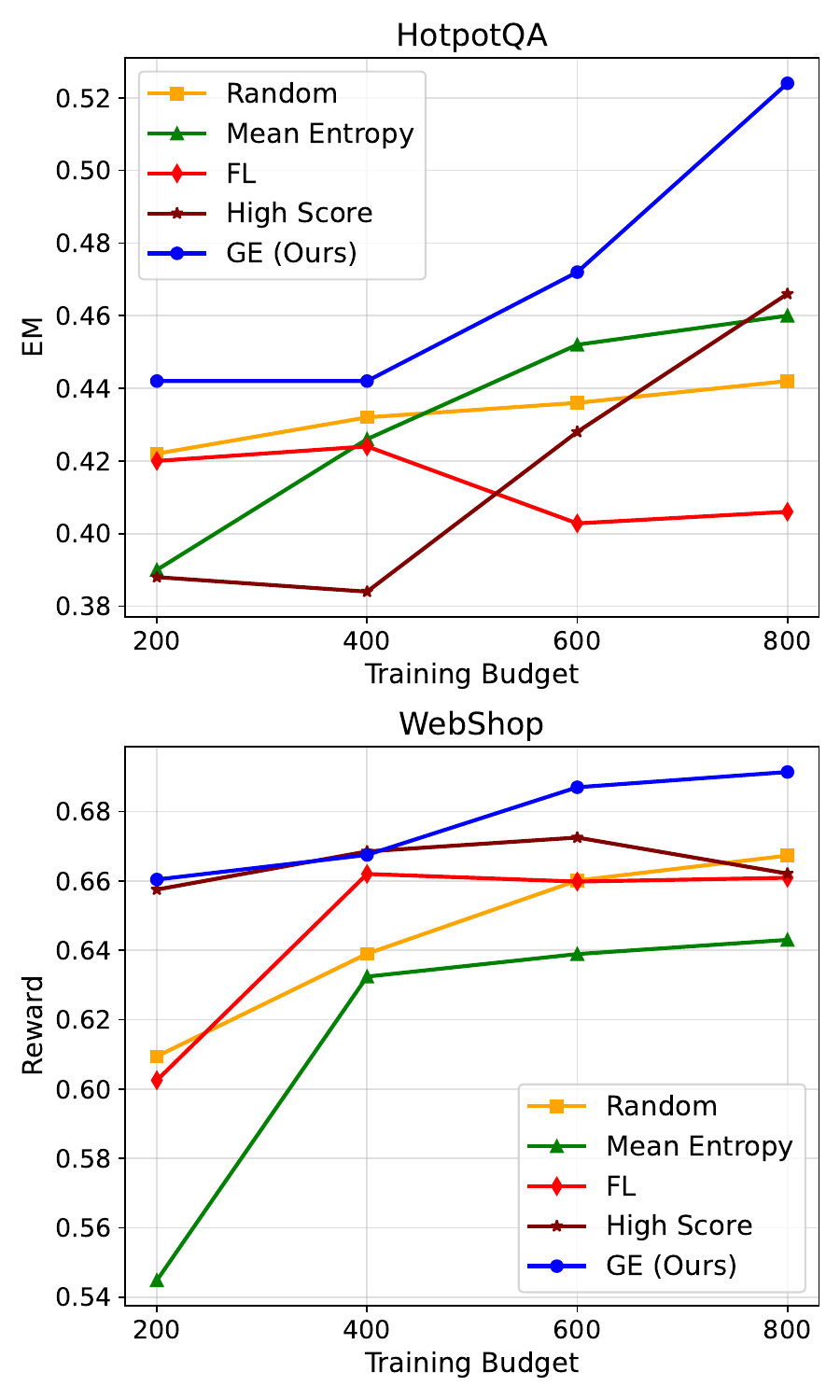}
        \caption{Comparison of different data selection strategies on various training budgets.}
        \label{fig:Lines}
    \end{minipage}
\end{figure*}
 
\textbf{Evaluation Metrics.}
For WebShop, $\text{reward} \in [0,1]$ measures how well the purchased item matches the text instruction, and SR measures the proportion of items that get reward=1. For HotpotQA, we employ two metrics: F1 and exact match (EM). Similar to SR, EM calculates the proportion of items whose F1=1. 

\textbf{Implementation details.} We invokes the OpenAI GPT4-4o (gpt-4o-2024-08-06) API. For all inference, we set temperature=0.7, top\_p=0.95, max\_length=512. For fine-tuning, we choose \texttt{LLAMA-3.1-8B-Instruct} (L-8B) and \texttt{Mistral-7B-Instruct-v0.3} (M-7B), training for four epochs with a learning rate of 5e-6 using 8 NVIDIA 80GB A100 GPUs. We use L-8B for the computation of GE score.


\subsection{Main Result}


\textbf{EDGE yield effective guidelines.}
$\text{EDGE}_\text{UG}$ surpasses baselines in both HotpotQA and WebShop,as shown in Table \ref{tab:maintab}, achieving improvements of 13.3\% and 13.9\%, respectively. 
The ExpeL autonomously summarizes guidelines, but its effectiveness is constrained by the LLM's limited environmental understanding. This shortcoming hinders ExpeL's ability to generate guidelines that demand a deeper comprehension of the environment or tools. For example, WebShop displays candidate products in a semantic similarity ranking, making the top-ranked products more likely to be target products. Due to ExpeL's lack of understanding of the search engine, it is unable to summarize related guidelines.

\textbf{GE-selected fine-tuning data outperform others.} We use same prompt to generate fine-tuning dataset for other data selection methods. Results in Table \ref{tab:maintab} show that GE outperforms them across the two datasets using L-8B and M-7B. Notably, ReAct w/ GE (L-8B) even surpassed  baselines that used GPT-4o. These findings indicate that the samples selected by GE are more challenging, and their solving trajectories integrate a greater depth of human expertise. 
FL focuses on selecting samples that are more semantically representative. Although it performs well on generation tasks, results indicate that it struggles with complex tasks that involve multi-turn interactions requiring reasoning and decision-making.
High Score, which filters and selects entirely accurate samples from labeled data, performs relatively well. Notably, not all samples selected by GE are labeled totally correctly. This means that despite having higher rewards, the data selected by the High Score still yields inferior results compared to GE.

\textbf{GE efficiently enhances fine-tuning.} We compared the performance of different data selection methods with training budget  k=[200, 400, 600, 800], as shown in Figure \ref{fig:Lines}. GE consistently outperforms the baselines across all training budgets, achieving more efficient integration of human expertise. Moreover, our method reduces training data usage by 50\% on WebShop and 25\% on HotpotQA, respectively, while still achieving better performance compared to the baselines.

\subsection{Analysis of EDGE}\label{sec:4-4}

\begin{figure*}[ht]
    \centering
    \includegraphics[page=1, width=1\textwidth]{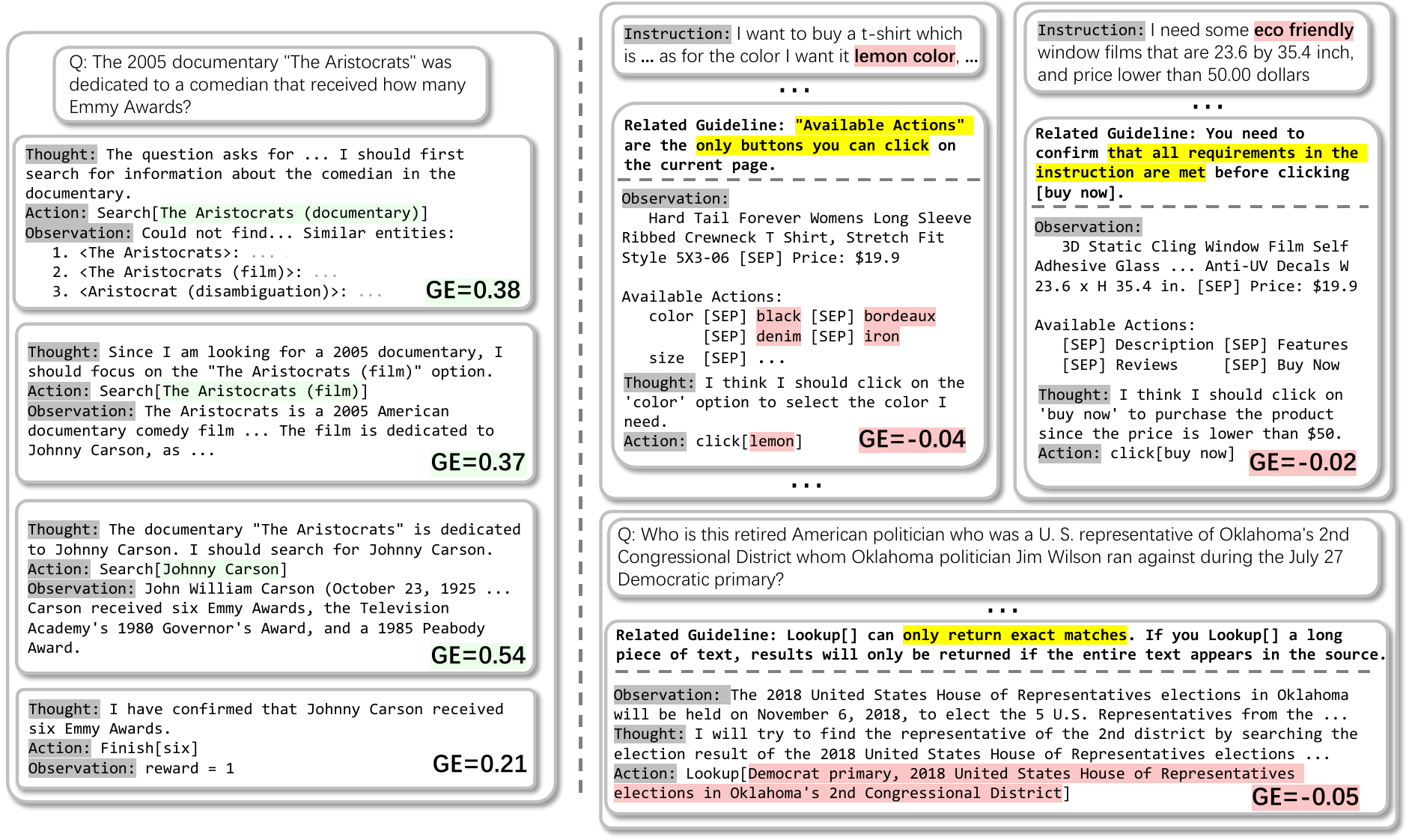}
    \caption{Examples of various actions' GE scores. The left side shows an example with a high GE score. The right side shows examples with lower GE scores on Webshop and HotpotQA. The red highlight indicates the reason for the lower GE value of the action.} 
    \label{fig:example}
\end{figure*}


\textbf{What kinds of samples have high/low GE score?} Through manual observation of the samples, we found that GE values tend to be higher when the environment is simple, and the guideline incorporates relevant human expertise. Conversely, GE values are lower in complex environments or when the guideline lacks relevant expertise. These observations align with our hypothesis. We selected a few samples with high/low GE score for illustration. On HotpotQA, Figure \ref{fig:example} (Left) is a sample with high GE score. The LLM followed the guideline, progressively searching for related entities based on the topic entity until the answer was found. Figure \ref{fig:example} (Right) demonstrates actions with low GE score. On WebShop, samples with complex environment caused the LLM to violate the guidelines, resulting in low GE scores. 
In the top left example, the extensive list of available actions on the product page, combined with relevant information, misled the LLM into click[lemon], which was a non-existent button. In the example at the top right corner, because the product met the other requirements in the instructions, the LLM hastily clicked the purchase button. The example below is a HotpotQA case, where this sample made the LLM use overly long search keywords.

\begin{figure}[ht]
    \centering
    \includegraphics[page=18, width=0.5\textwidth]{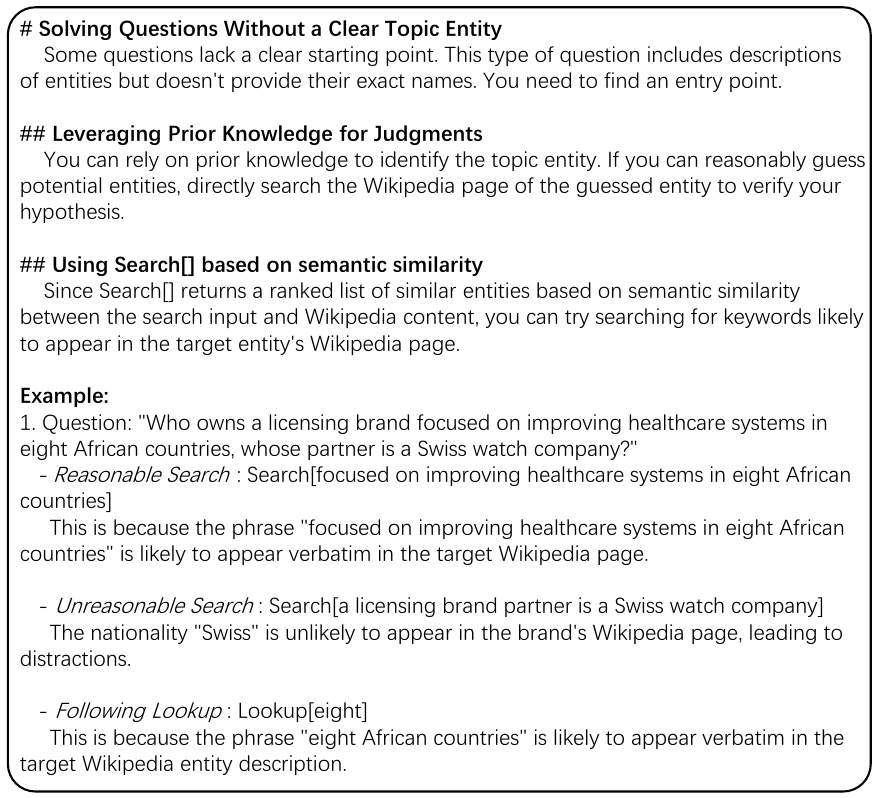}
    \caption{Example of a peice of our guideline on HotpotQA} 
    \label{fig:example_guideline}
\end{figure}

\textbf{What guidelines have we summarized?} Based on the observation of low GE score samples, we have listed the following representative guidelines in order of their necessity within the samples. Figure \ref{fig:example_guideline} shows the format of our guideline. For WebShop: 
\begin{itemize}
    \item \textit{Click high-ranked products first.} A high ranking indicates that the semantics of the product's options or features are more similar to the search content. Even if the titles of high-ranked products sometimes do not fully meet the required criteria, it is still necessary to click on them.

    \item \textit{What is a product title.} This guideline introduces that some attributes like color, flavor, etc., will not appear in the product title. LLMs often hope to find products whose titles perfectly match the instructions.

    \item \textit{Buying a similar product is better than buying nothing.} The LLM needs to complete the purchase of a product within a limited number of interaction rounds. Therefore, before the rounds are exhausted, the LLM needs to balance the trade-off and compromise to buy a product that is not perfectly aligned when necessary.

\end{itemize}

For HotpotQA:
\begin{itemize}

    \item \textit{How to use `Lookup'.} Despite being informed that "the Lookup[] only supports exact matching," the LLM still tends to search for longer keywords. This guideline details that the LLM should search for the shortest possible word that are likely to appear in the original text to match more search results.

    \item \textit{Solving questions without a clear topic entity} When the topic entity cannot be found, LLM can try leverage its prior knowledge to leverage its prior knowledge to make a reasonable inference or use Search[] based on semantic similarity.

    \item \textit{Understanding the Question.} This guideline provides a detailed explanation on how to understand and deconstruct complex problems. In complex scenarios, if certain representative requirements are met(e.g., the 20th President of the United States), the LLM can respond directly without confirming other constraints.

\end{itemize}

\begin{table*}[t]
    \centering
    \begin{tabular}{@{}lrrrrrrrrc@{}}
        \toprule
        \multirow{2}{*}{\textbf{Method}} & \multicolumn{4}{c}{\textbf{Avg. Interaction Turns}} & \multicolumn{4}{c}{\textbf{Reward}}  &  \multirow{2}{*}{\textbf{LLM Reward}} \\ 
        \cmidrule(lr){2-5} \cmidrule(lr){6-9} 
                                 & k=200 & k=400 & k=600 & k=800 & k=200 & k=400 & k=600 & k=800 & \\ \midrule
        FL                     & 4.50   & 4.53   & 4.45   & 4.43   & 80.85 & 79.78 & 77.86 & 77.68 & 46.06\\
        Random                  & 4.97   & 4.84   & 4.75   & 4.87   & 77.56 & 79.21 & 79.81 & 77.69 & 46.11\\
        Mean Entropy            & 4.82   & 4.92   & 4.89   & 4.96   & 72.17 & 72.97 & 73.18 & 71.10 & 47.00\\
        High Score              & 4.77   & 4.60   & 4.65   & 4.64   & 100.00 & 100.00 & 100.00 & 100.00 & 49.19\\
        \midrule
        GE (Ours)                     & 6.40 & 6.38 & 6.43 & 6.39 & 68.06 & 70.84 & 71.09 & 70.27 & 55.47\\ \bottomrule
    \end{tabular}
    \caption{Statistics of the annotated data by different data selection methods.}

    \label{tab:Statistics}
\end{table*}

\textbf{Dose filtering reward=1 trajectories truly lead to high quality?} The quality of the fine-tuned dataset directly impacts the performance of the fine-tuned LLM. Intuitively, annotated samples with  higher rewards suggest higher quality. Thus, existing methods often filter fully correct samples (reward = 1) from many annotated examples, known as High Score. However, does low reward always indicate low quality? To answer this question, we analyzed the annotated samples selected by different  methods, as shown in Table \ref{tab:Statistics}. Compared to Random, High Score has lower average interaction turns, which typically indicates that the samples are simpler and easier. Combined with the observation from Table \ref{tab:proportion} that High Score is least likely to select hard questions, we can conclude that High Score includes more simple samples during filtering. This explains why High Score fails to achieve the best performance despite all data reward=1. Notably, despite the lowest annotated data rewards of GE, it achieves the best performance. This is because more challenging samples can better leverage the advantages of human experience from the guidelines. This suggests that data containing more ``attempts at challenging problems"  represents higher quality for fine-tuning, even if it is not fully correct.


\subsection{Effective Analysis in HotpotQA}

To investigate whether EDGE expanded the range of solvable problems, we analyzed the distribution of question difficulty levels in HotpotQA (easy, medium and hard). Table \ref{tab:proportion} presents the proportion of each difficulty level within the subsets selected by different methods. Among the various approaches, only GE selected a higher proportion of hard questions. With more out-of-guideline questions selected for annotating, GE achieved slight advantages on easy and medium questions, and significantly outperformed the baselines on hard questions, as shown in Table \ref{tab:Performance-on-level}. Consequently, GE effectively broadened the scope of solvable problems by focusing on out-of-guideline questions.
\begin{table}[t]
    \centering
    \begin{tabular}{lccc}
        \toprule
        Method & \textbf{Easy} & \textbf{Medium} & \textbf{Hard} \\
        \midrule
        Random & 0\% & 0\% & 0\% \\
        Mean Entropy & +1.08\% & +1.13\% & -2.21\% \\
       High Score & -0.92\% & +4.13\% & -3.21\% \\
        FL & +0.71\% & +0.00\% & -0.71\% \\
        \midrule
        GE (Ours) & -1.67\% & -1.50\% & +3.17\% \\
        \bottomrule
    \end{tabular}
    \caption{The proportion of different difficulty levels across various methods. The percentages indicate the change in proportion compared to the original distribution of difficulty levels}
    \label{tab:proportion}
\end{table}

\begin{table}[t]
    \centering
    \begin{tabular}{lccc}
        \toprule
        \textbf{Method} & \textbf{Easy} & \textbf{Medium} & \textbf{Hard} \\
        \midrule
        ReAct &-&-&- \\
        \quad w/ Random & 65.29 & 62.92 & 56.02 \\
        \quad w/ Mean Entropy  & 62.12 & 62.29 & 56.13 \\
        \quad w/ High Score & 66.68 & 65.11 & 57.66 \\
        \quad w/ FL & 63.19 & 61.82 & 52.38 \\
        \midrule
        \quad w/ GE (Ours) & 68.64 & 66.98 & 66.15 \\
        \bottomrule
    \end{tabular}
    \caption{Performance comparison on different difficulty levels}
    \label{tab:Performance-on-level}
\end{table}

\section{Conclusion}
We propose GE metric, which can effectively identifies the most informative samples without requiring a golden answer. Selecting samples with low GE scores enhances the efficiency and outcomes of prompt engineering and fine-tuning processes for LLMs. Extensive experiments demonstrate the effectiveness of our method, and we finally provides a fresh perspective on the data quality of LLM-agent fine-tuning.


\bibliographystyle{named}
\bibliography{ijcai24,ref/agent_plan.bib}


\end{document}